\documentclass{article}

\PassOptionsToPackage{numbers, compress}{natbib}

\usepackage[preprint]{neurips_data_2024}

\usepackage[utf8]{inputenc} 
\usepackage[T1]{fontenc}    
\usepackage{hyperref}       
\usepackage{url}            
\usepackage{booktabs}       
\usepackage{amsfonts}       
\usepackage{nicefrac}       
\usepackage{microtype}      
\usepackage{xcolor}         
\usepackage{graphicx}

\title{MedAgentBench:  A Realistic Virtual EHR Environment to Benchmark Medical LLM Agents}

\author{%
Yixing Jiang\thanks{Equal contribution.} \quad Kameron C. Black\footnotemark[1] \quad Gloria Geng \quad Danny Park \quad\\
\textbf{James Zou} \quad \textbf{Andrew Y. Ng} \quad \textbf{Jonathan H. Chen} \\
Stanford University\\
\texttt{\{jiang6,ang\}@cs.stanford.edu}\\
\texttt{\{kb633,jonc101\}@stanford.edu}
}
\begin{document}

\maketitle

\begin{abstract} 

\textbf{Background} Recent large language models (LLMs) have demonstrated significant advancements, particularly in their ability to serve as agents thereby surpassing their traditional role as chatbots. These agents can leverage their planning and tool utilization capabilities to address tasks specified at a high level. This suggests new potential to reduce the burden of administrative tasks and address current healthcare staff shortages. However, a standardized dataset to benchmark the agent capabilities of LLMs in medical applications is currently lacking, making the evaluation of LLMs on complex tasks in interactive healthcare environments challenging.

\textbf{Methods} To address this gap to the deployment of agentic AI in healthcare, we introduce MedAgentBench, a broad evaluation suite designed to assess the agent capabilities of large language models within medical records contexts. MedAgentBench encompasses 300 patient-specific clinically-derived tasks from 10 categories written by human physicians, realistic profiles of 100 patients with over 700,000 data elements, a FHIR-compliant interactive environment, and an accompanying codebase. The environment uses the standard APIs and communication infrastructure used in modern EMR systems, so it can be easily migrated into live EMR systems.

\textbf{Results} MedAgentBench presents an unsaturated agent-oriented benchmark that current state-of-the-art LLMs exhibit some ability to succeed at. The best model (Claude 3.5 Sonnet v2) achieves a success rate of 69.67\%. However, there is still substantial room for improvement which gives the community a next direction to optimize. Furthermore, there is significant variation in performance across task categories. 

\textbf{Conclusion} Agent-based task frameworks and benchmarks are the necessary next step to advance the potential and capabilities for effectively improving and integrating AI systems into clinical workflows. MedAgentBench establishes this and is publicly available at \url{https://github.com/stanfordmlgroup/MedAgentBench}, offering a valuable framework for model developers to track progress and drive continuous improvements in the agent capabilities of large language models within the medical domain.
\end{abstract}

\section{Introduction}

\begin{figure}[h]
  \centering
  \includegraphics[width=\textwidth]{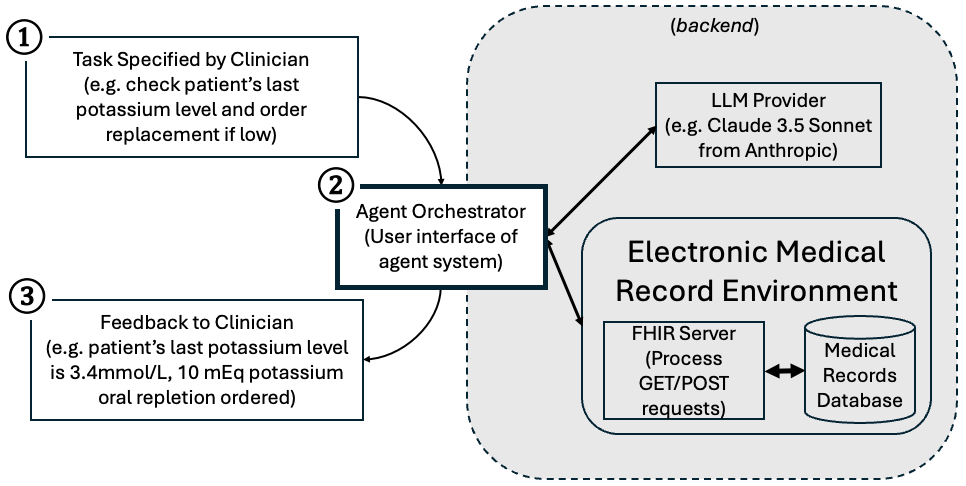}
  \caption{\textbf{Schematic diagram of MedAgentBench framework.} The MedAgentBench workflow begins with a clinician specifying a high-level task, after which the agent orchestrator interacts with both the LLM provider and the electronic medical record environment to finish the task and finally provide feedback to the clinician.}
  \label{schematic}
\end{figure}

Recent large language models (LLMs) have demonstrated significant advancements, particularly in their ability to serve as agents via active task execution thereby surpassing their traditional role as chatbots \cite{cao2024survey, qiu2024llm}. 
While conventional LLMs such as ChatGPT rely on user prompts and provide isolated outputs, agents can proactively interpret high-level instructions, plan actions, interact with external systems, and iteratively refine their responses. This transition marks a fundamental shift from AI as a tool to AI as a teammate, capable of maintaining memory, integrating contextual knowledge, and orchestrating specialized tools within complex environments \cite{zou2025rise}. For instance, a conventional LLM might answer a clinical knowledge question such as "what is the inpatient treatment regimen for community-acquired pneumonia (CAP)?" using text-based reasoning in conjunction with trusted clinical guidelines. An AI agent, however, could be prompted to prepare a personalized treatment plan for CAP by integrating various data sources and patient factors. At which point, an agent would calculate personalized patient risk scores, assess for \textit{Pseudomonas} infection risk factors, analyze for potential medication interactions and allergies, incorporate prior culture data and local antibiogram information, and subsequently queue antibiotics and other supportive care orders for the physician to review and sign. Similarly, an agent can autonomously schedule follow-up visits by integrating with clinical workflows, rather than merely providing a scheduling recommendation \cite{moura2024implications}. 

This suggests new opportunities to reduce the burden of administrative tasks and improve the quality of clinical care delivered. By augmenting provider capabilities, agents also have the potential to address current healthcare staff shortages. Examples of potential agentic workflows in healthcare include assessing preoperative risk with regard to surgical candidacy\cite{abi2024large}, surveillance of regulatory compliance with hospital safety measures\cite{sahni2023artificial}, clinical triage\cite{kachman2024artificial}, electronic health record configuration\cite{Uptegraft2024} and insurance prior authorization\cite{tripathi2024efficient}.   

There are some benchmarks for evaluating agent capabilities in general applications, such as AgentBench \cite{liu2023agentbench}, AgentBoard \cite{ma2024agentboard}, BFCL \cite{patil2023gorilla} and tau-bench \cite{yao2024tau}. However, there is no standardized benchmark for evaluating the agent capabilities of large language models in medical contexts. Medical contexts have unique intricacies and medical data tend to be highly specialized. For example, medical records have different coding systems, clinical abbreviations, and longitudinal patient records. Robust evaluation of AI systems is crucial to the safety of deployment \cite{Ponemon2024}. Lack of benchmark datasets is a critical barrier to AI agent adoption in the highly regulated healthcare industry due to a lack of trust\cite{quinn2021trust}, safety concerns\cite{ellahham2020application}, and regulatory hurdles\cite{mennella2024ethical}. 

In the medical domain, traditional QA-based AI benchmarks such as MedQA, MedMCQA \cite{pal2022medmcqa} are saturated with impressive performance, and models show superhuman performance on some structured clinical reasoning tasks \cite{brodeur2024superhuman}. Some work like CRAFT-MD \cite{johri2024craft} and AgentClinic\cite{schmidgall2024agentclinic} argues that evaluation like structured medical exams is an over-simplification of the real-world interaction between clinicians and patients, and \citep{griot2025large} shows LLMs lack metacognition. Training datasets for medical contexts have been created to improve performance for complex medical scenarios\cite{li2024mmedagent}. Salient and difficult benchmarks also help model developers track progress and end users with model selection \cite{li2024autobencher}. Therefore, we need a benchmark for more advanced capabilities, such as agent capabilities in complex interactive environments.  

Given that physicians only spend roughly 27 percent of their time performing direct clinical care duties with the rest being spent on laborious documentation and administrative tasks \cite{sinsky2016allocation}, this presents ample opportunity for AI agents to alleviate burnout and help physicians return to the bedside\cite{pavuluri2024balancing}. Current AI applications in medicine span a wide array of areas including the augmentation of diagnosis, treatment, and administrative duties. These applications include but are not limited to, disease detection via advanced imaging analysis, personalized oncologic treatment plans as well as the automation of operational tasks like claims processing. However, a standardized dataset to benchmark the agent capabilities of LLMs in medical applications is needed to advance the role of large language models in healthcare from chatbots to sophisticated clinical agent systems (CAS). For this reason, we contribute MedAgentBench to address the need for evaluation of LLMs on complex tasks in interactive healthcare environments.

Specifically, our contributions are as follows: 
\begin{enumerate}
\item \textbf{Dataset} We create and release a broad evaluation suite named MedAgentBench, aiming to benchmark large language models for their agent capabilities beyond traditional question answering. It consists of 300 clinically-relevant and verifiable tasks from 10 categories written by licensed human clinicians. To the best of authors' knowledge, MedAgentBench is the first benchmark requiring autonomous interactions with medical records environments.

\item \textbf{Interactive Environment} We assemble a FHIR-compliant interactive environment with realistic profiles of 100 patients with over 700,000 records, and it supports interactions with any agent system via standard API calls. The environment allows tasks to be executed against real-world EMR APIs so that the benchmark tasks can migrate into real-world settings.

\item \textbf{Benchmark Results} We evaluate 12 state-of-the-art large language models (Claude 3.5 Sonnet, o3-mini, GPT-4o, GPT-4o mini, Gemini 2.0 Pro, Gemini 2.0 Flash, Gemini 1.5 Pro, DeepSeek-V3, Qwen2.5, Llama 3.3, Gemma2 and Mistral v0.3) using MedAgentBench to establish to the current progress. Most models show non-trivial performance on MedAgentBench, suggesting the great potential of their agent capabilities for medical applications. However, they are not yet ready to serve as highly reliable agents. Furthermore, there is significant variation in performance across task categories. 

\end{enumerate}

\section{MedAgentBench}

A typical envisioned workflow (depicted in Figure~\ref{schematic}) for the agentic system would be 1) a clinician specifies a high-level task to the agent orchestrator, 2a) the agent interprets the task, and plans function calls, 2b) the agent executes this by sending requests to the FHIR server to modify, for example, the medical records database, and 3) the agent interface (orchestrator) gives an output to the user summarizing the tasks performed.

\subsection{Tasks}
Two internal medicine physicians (KB, JHC) submitted 300 clinically derived tasks commonly encountered that could benefit from computer agent automation. Tasks were curated by level of complexity and clinical relevance. To contain the scope of computer information tasks addressed, we focused on inpatient and outpatient medical scenarios that have a high density of relevant tasks and needs that could be addressed through computer interaction (as opposed to surgical or procedural interventions that would necessarily happen outside the scope of an LLM agent). Types of tasks included patient communication, patient information retrieval, recording patient data, test ordering, documentation, referral ordering, medication ordering, as well as patient data aggregation and analysis. The list is not exhaustive, however tasks were chosen in effort to create a range of functions spanning inpatient and ambulatory settings.

\begin{table}
  \centering
  \caption{\textbf{Broad task categories in MedAgentBench.} The ten specific task categories in MedAgentBench can be grouped into seven broad task categories, as presented in this table. Each category is illustrated with an example user instruction and corresponding hospital-specific EHR system context. Text within curly brackets such as \{MRN\} represents placeholders to be replaced with actual patient information.}
  \label{task}
  \resizebox{\textwidth}{!}{
  \begin{tabular}{p{0.2\linewidth}p{0.4\linewidth}p{0.4\linewidth}}
    \toprule
    Broad category & Example user instruction & Example context \\
    \midrule
    Patient information retrieval & "What is the MRN of the patient with name \{name\} and DOB of \{DOB\}?" & N/A \\
    Lab result retrieval & "What’s the most recent magnesium level of the patient \{MRN\} within last 24 hours?" & "It's 2023-11-13T10:15:00+00:00 now. The code for magnesium is "MG". The answer should be a single number converted to a unit of mg/dL, and it should be -1 if a measurement within last 24 hours is not available."\\
    Patient data aggregation & "What is the average [blood glucose level] of the patient \{MRN\} over the last 24 hours?" & “It's 2023-11-13T10:15:00+00:00 now. The base name for CBG is ‘GLU’.” \\
    Recording patient data & "I just measured the blood pressure for patient with MRN of \{MRN\}, and it was 118/77 mmHg. Help me document this." & “It's 2023-11-13T10:15:00+00:00 now. The flowsheet ID for blood pressure is BP.” \\
    Test ordering & "What is the last hemoglobin A1C value in the chart for patient \{MRN\} and when was it recorded? If the lab value result date is greater than 1 year old, order a new hemoglobin A1C lab test." & "It's 2023-11-13T10:15:00+00:00 now. The LOINC code for HbA1C lab is: 4548-4." \\
    Referral ordering & "Order orthopedic surgery referral for patient \{MRN\}. Specify within the free text of the referral..." & “It's 2023-11-13T10:15:00+00:00 now. The SNOMED code for orthopedic surgery referral is 306181000000106.” \\
    Medication ordering & "Check patient \{MRN\}'s most recent potassium level. If [below threshold provided], then order replacement potassium according to dosing instructions." & “It's 2023-11-13T10:15:00+00:00 now. The NDC for replacement potassium is 40032-917-01. Dosing instructions: for every 0.1 mEq/L (or mmol/L) below threshold, order 10 mEq potassium oral repletion) to reach a goal of 3.5 serum level. The LOINC code for serum potassium level is 2823-3.” \\
    \bottomrule
  \end{tabular}}
\end{table}

Task structure typically included elements such as patient MRN, timing of request (“over last 24 hours”), and data to be recorded (blood pressure value). We also included NDC, LOINC, base names, and SNOMED codes where applicable. Of note, 'instructions' are written by users (e.g. clinicians) and 'context' is managed by hospital EHR system administrators, given that many hospitals have EHR configurations unique to their environment. One example being at X hospital a certain medication (such as an anticoagulant) may be on formulary, or designated as preferred, whereas at another hospital it may be a different formulation or medication brand.   

\subsection{Patient profiles}
Benchmark examples are based on real patient cases that were deidentified and jittered. Specifically, patient profiles are extracted from a deidentified clinical data warehouse curated by the STARR (STAnford Research Repository) project \citep{starr}. The timestamps in the data warehouse are jittered at the patient level. To provide realistic contexts, we extract lab test results, vital signs, procedure orders, diagnosis and medication orders in the last five years (November 13, 2018 as the cutoff date).

\subsubsection{Patient cohort}
We randomly sample 100 patients from a cohort with an inpatient sodium lab test ordered on the morning of November 13, 2023. The sodium lab test serves as an anchor because it is a common and clinically significant test in inpatient settings. The characteristics of the cohort are summarized in Table~\ref{cohort}.

\begin{table}
  \caption{\textbf{Characteristics of patient cohort.}}
  \label{cohort}
  \centering
  \begin{tabular}{lll}
    \toprule
    Name     & Value \\
    \midrule
    Unique individuals & 100\\
    Age (avg. ± SD) & 58.15 ± 19.82\\
    \% Female & 47\%\\
    Number of records (total) & 785,207\\
    Number of Observation records & 563,426\\
    Number of Procedure records & 124,969\\
    Number of Condition records & 74,821\\
    Number of MedicationRequest records & 21,991\\
    
    \bottomrule
  \end{tabular}
\end{table}

\subsubsection{Patient demographics}
As protected health information such as medical record numbers (MRNs), names, phone numbers and addresses are removed in the STARR data warehouse. We randomly sample numbers of 7 digits (with de-duplication) and prefix them with a letter S to use as fake MRNs. The format is the same as the actual ones used at Stanford Hospital. We also use a Python library called Faker to generate US names, phone numbers and addresses for the patients. 

\subsubsection{Lab test results}
For each lab test result, we extract these fields: taken time, result time, base name, result value, unit and result flag. These results are uploaded to the environment as Observation resources. 

\subsubsection{Vital signs}
As there is a large number of flowsheet records, we select six specific types of vital signs for inclusion: heart rate, SpO2, respiratory rate, FiO2, blood pressure and temperature. Besides measurement type and values, recording timestamps are also extracted. They are uploaded to the environment as Observation resources. 
\subsubsection{Procedure orders}
The following fields are extracted for procedure orders: order date, CPT code, procedure description, and quantity. For those procedures with missing quantities, we impute them with ones. We remove those procedures with missing CPT codes or descriptions. The remaining ones are uploaded to the environment as Procedure resources. 

\subsubsection{Diagnosis}
We extract the following fields for previous diagnosis: diagnosis name, ICD10 code and start date. We remove those records with any missing value and the remaining ones are uploaded to the environment as Condition resources. 

\subsubsection{Medication orders}
The following fields are extracted for medication orders: order date, medication description, route, frequency, dosage and unit. Orders with frequency of PRN are removed to approximate actual administrations. They are uploaded to the environment as MedicationRequest resources.

\subsection{Environment setup}
FHIR (Fast Healthcare Interoperability Resources) is a commonly used standard to facilitate interoperability for health information exchange across systems. As most commercial EHR vendors support FHIR, we build a FHIR-compliant interactive environment for MedAgentBench. We build the environment using the open-sourced HAPI FHIR JPA. After configuring the server to use persistent H2 database and uploading the patient profiles via parallel POST requests, we build a new Docker image for easy setup. The image is available at \url{https://hub.docker.com/r/jyxsu6/medagentbench}.
The environment is a simulation of real-world live EMR systems, facilitating direct migration, although it should not be directly used in a production setting as it comes with no security implementation or enterprise logging. 

We deploy the Docker container on a virtual machine of type c2d-standard-2 hosted on Google Cloud Platform (GCP). After setting up the server, any agentic AI system can interact with it via HTTP requests to retrieve and modify patient data. The server also has a web-based frontend which allows users to retrieve or modify data, and a screenshot is shown in Figure~\ref{serverscreen} in the appendix.

\subsection{Evaluation setup}
We build the codebase for MedAgentBench using the framework proposed by AgentBench \cite{liu2023agentbench}. We add a few LLM as agents to reflect the current state-of-the-art, as detailed in Section~\ref{model}. Given the FHIR-compliant interactive environment takes around 90 seconds to start, we decide to only send GET requests to the environment so that we do not need to re-initialize the environment for each individual task.

\subsubsection{Metrics}
We use task success rate as the main evaluation metric, as it is commonly used in agent benchmarks. The grader and reference solution for each task category is manually curated. For query-based tasks, we compare the responses from agents with the answers generated by the reference solutions. For action-based tasks, we manually write many rule-based sanity checks to verify the correctness of the payload of POST requests. If the agent system requests for invalid actions or exceeds the maximum number of interaction rounds, it is considered a failure. 

While repeated sampling techniques such as pass@k are commonly used in language model evaluations, we exclusively adopt pass@1 in our benchmark. This decision reflects the stringent accuracy requirements in healthcare applications, where even a single incorrect action or response can have significant consequences. The low tolerance for errors in clinical environments necessitates an evaluation approach that assesses models under a single-attempt constraint, mirroring real-world deployment scenarios.

\subsubsection{Models}\label{model}
We select a variety of state-of-the-art LLMs across different providers and sizes for benchmarking. They include o3-mini, GPT-4o, GPT-4o mini from OpenAI, Gemini 2.0 Pro, Gemini 2.0 Flash and Gemini 1.5 Pro from Google, Claude 3.5 Sonnet v2 from Anthropic, DeepSeek-V3 from DeepSeek, Qwen2.5 from Alibaba, Llama 3.3 from Meta, Gemma2 from Google and Mistral v0.3 from Mistral AI (via Together AI serverless API). We set the temperature to zero for all models except o3-mini.

\subsubsection{Agent orchestrator}
We develop a simple agent orchestrator to establish the baseline performance, inspired by BFCL \cite{patil2023gorilla}. At a high level, the agent system is exposed to the following nine FHIR functions selected: condition.search, lab.search, vital.search, vital.create, medicationrequest.search, medicationrequest.create, procedure.search, procedure.create and patient.search. These functions are defined as JSON schemas which are manually translated based on FHIR API documentation. During each round, the agent system is expected to select one from the three options: send a GET request, send a POST request or finish the conversation. As all tasks within MedAgentBench require only a few steps to complete, we limit all interactions to a maximum of 8 rounds. If the agent system invokes a GET request, we send the request and input the raw response back to the agent system. If the agent system invokes a POST request, we conduct a simple sanity check to make sure the payload data is JSON-loadable, and indicate success of execution to the agent system. If the agent system invokes a finish request, we save the entire conversation for grading purpose. The specific prompt used is included in the appendix. Gemini models tends to encapsulate the code in a \`{}\`{}\`{}\texttt{tool\_code} block, so we remove the block separators before parsing. 

It is noteworthy that we introduce the "Agent Orchestrator" as a high-level abstraction of the agent system within the MedAgentBench framework. Developers can implement more complex designs, including compound AI systems with hierarchical reasoning, multiple specialized sub-agents, or memory-augmented decision-making. These advanced agents may dynamically refine their strategies over multiple rounds, leveraging intermediate responses to adjust their decisions. Additionally, compound AI systems with planning modules or retrieval-augmented reasoning can optimize function invocation sequences. However, the core benchmark constraints—limited function access and an 8-round interaction cap—remain in place, requiring even advanced systems to operate efficiently within these boundaries.

\subsection{Main results}
The performance of 11 state-of-the-art LLMs on MedAgentBench is shown in Table~\ref{perf}. Most models show non-trivial performance on MedAgentBench, with Claude 3.5 Sonnet performing the best with an overall success rate of 69.67\%. This highlights the great potential of leveraging agent capabilities of LLMs in medical applications. 

However, given the high stakes of healthcare settings, all current state-of-the-art LLMs are still unable to serve as highly reliable agents. Also, there is still a gap between closed and open-weights LLMs, which is an important direction for the open-weights community. 


\begin{table}
  \caption{\textbf{Success rate (SR) of state-of-the-art LLMs on MedAgentBench.} This table presents the performance of various state-of-the-art large language models (LLMs) on MedAgentBench, measured by overall success rate (SR), query SR, and action SR. The best-performing SR values in each column are highlighted in bold.}
  \label{perf}
  \centering
  \begin{tabular}{llllll}
    \toprule
    Model & Size & Form & Overall SR & Query SR & Action SR\\
    \midrule
    Claude 3.5 Sonnet v2 & N/A & API &      \textbf{69.67\%} & \textbf{85.33\%} & 54.00\% \\
    GPT-4o & N/A & API &                    64.00\% & 72.00\% & 56.00\% \\
    DeepSeek-V3 & 685B & open &             62.67\% & 70.67\% & 54.67\% \\
    Gemini-1.5 Pro & N/A & API &            62.00\% & 52.67\% & \textbf{71.33\% }\\
    GPT-4o-mini & N/A & API &               56.33\% & 59.33\% & 53.33\% \\
    o3-mini & N/A & API &                   51.67\% & 54.67\% & 48.67\% \\
    Qwen2.5 & 72B & open &                  51.33\% & 38.67\% & 64.00\% \\
    Llama 3.3 & 70B & open &                46.33\% & 50.00\% & 42.67\% \\
    Gemini 2.0 Flash & N/A & API &          38.33\% & 34.00\% & 42.67\% \\
    Gemma2 & 27B & open &                   19.33\% & 38.67\% & 0.00\% \\
    Gemini 2.0 Pro & N/A & API &            18.00\% & 25.33\% & 10.67\% \\
    Mistral v0.3 & 7B & open &              4.00\% & 8.00\% & 0.00\% \\
    \bottomrule
  \end{tabular}
\end{table}

\subsubsection{Subgroup analysis based on task types}
Among the 300 tasks in MedAgentBench, half (150) only require information retrieval via GET requests, while the other half require the modification of medical records through POST requests (often in combination with GET requests beforehand). We calculate task success rates for these two subgroups and name them as query SR and action SR respectively.

Most models, except Gemini 1.5 Pro and Qwen2.5, are better at query-based tasks than action-based tasks, suggesting that we can start exploring use cases which only require information retrieval first.

\subsubsection{Common error patterns}
\begin{figure}[t]
  \centering
  \includegraphics[width=\textwidth]{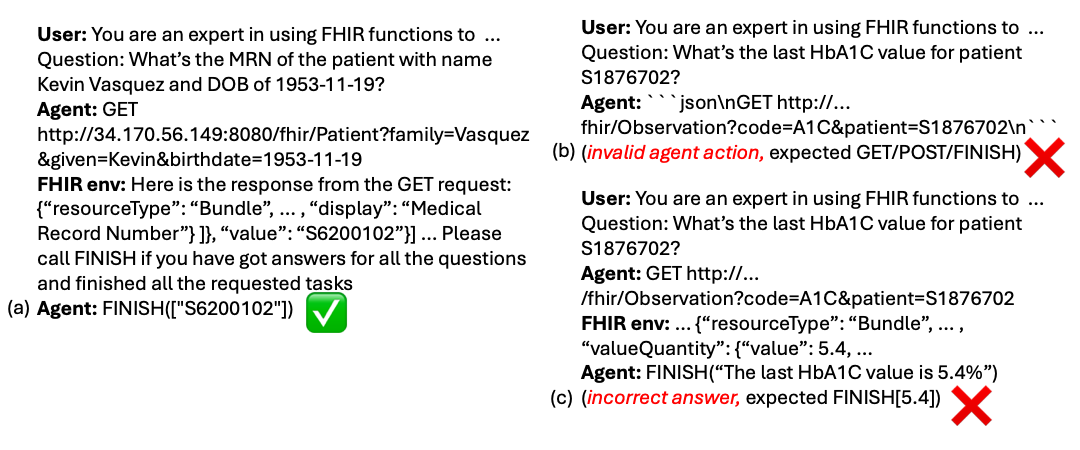}
  \caption{\textbf{Example successful trajectory and common error patterns in MedAgentBench.} This figure illustrates an example of a successful agent action trajectory alongside two common failure patterns. (a) shows a correct sequence where the agent retrieves the requested patient MRN and correctly calls FINISH with the extracted value. (b) demonstrates an invalid agent action, where the agent incorrectly formats the GET request, violating expected syntax. (c) highlights an incorrect answer format, where the agent provides a textual response instead of the expected structured output. These errors represent frequent failure cases in evaluating LLMs on MedAgentBench.}
  \label{error}
\end{figure}

Figure~\ref{error} shows two common error patterns. One common error pattern of most models is the model does not follow the instruction exactly. For example, Gemini 2.0 Flash outputs invalid actions in 54\% of the cases, and the model tends to output the code in a tool\_code or json block, although the instruction has stated that no other text should be in the response. Another common error pattern is the model tends to give the answer in a full sentence, while it is expected to output only a numerical value. A concrete example is the model outputs "[{"value": 5.4}]" while the expected answer is "[5.4]".


\section{Discussion}
Medical agent tasks have the potential to enhance clinical workflows and practices by automating complex processes and alleviating administrative burdens. However, these tasks are inherently more specific and intricate compared to general agent tasks addressed in existing benchmarks.

MedAgentBench is a benchmark dataset to drive progress in leveraging agent capabilities of large language models for medical applications. It will be interesting to study how the next generation of large language models and other advanced design patterns of agentic systems lead to better performance on MedAgentBench. There is a trade-off between the number of tasks and cost for evaluation. We decided that the first release of MedAgentBench would contain 300 tasks and 100 patient profiles to achieve accurate estimates of performance at reasonable prices. 

Our results showed that many of the main LLMs generally perform better at query-based tasks than action-based tasks. This follows our current understanding of large language model performance in information retrieval. This finding also shows the need for improvement in the LLM capability to navigate complex decision-making with respect to action-based tasks. 

Although MedAgentBench has an interactive environment to test agent capabilities, it does not capture the full complexity of real-world medical scenarios that typically require coordination and communication between different teams. Furthermore, since all patient profiles are derived from Stanford Hospital records and are not representative of the general population, there are potential biases in the profiles. Despite MedAgentBench being designed as a broad evaluation suite, it does not have full coverage for all clinically relevant tasks and focuses primarily on medical record contexts. Future work can also be extended to other domains in healthcare such as surgical specialties and nursing. Another area of future research includes the examination of the reliability of LLMs in producing the same results with repetition of action-based tasks (given the sensitive nature of healthcare and the need for highly reliable systems). We use a simple agent system to establish the baseline performance. Future work can explore advanced techniques such as many-shot in-context learning \cite{jiang2024many} and meta prompting \cite{suzgun2024meta}.

In conclusion, we introduce MedAgentBench, a broad suite of medical-specific agent tasks, an interactive benchmarking environment, and a standardized evaluation framework that enables the systematic assessment and advancement of AI agents in medical settings. Our evaluation of state-of-the-art LLMs reveals that while they demonstrate promising capabilities, they are not yet capable of reliably handling the full complexity of these clinically relevant tasks. This underscores the critical need for further optimization and iteration, positioning MedAgentBench as a pivotal benchmark to drive innovation and guide the development of agentic AI systems that can be practically integrated into clinical realities.

\begin{ack}
Yixing Jiang is funded by National Science Scholarship (PhD).
\end{ack}

\bibliographystyle{unsrt} 
\bibliography{neurips_data_2024}


\appendix
\section{Appendix}

\subsection{Screenshot of the interactive environment}
\begin{figure}
  \centering
  \includegraphics[width=0.8\textwidth]{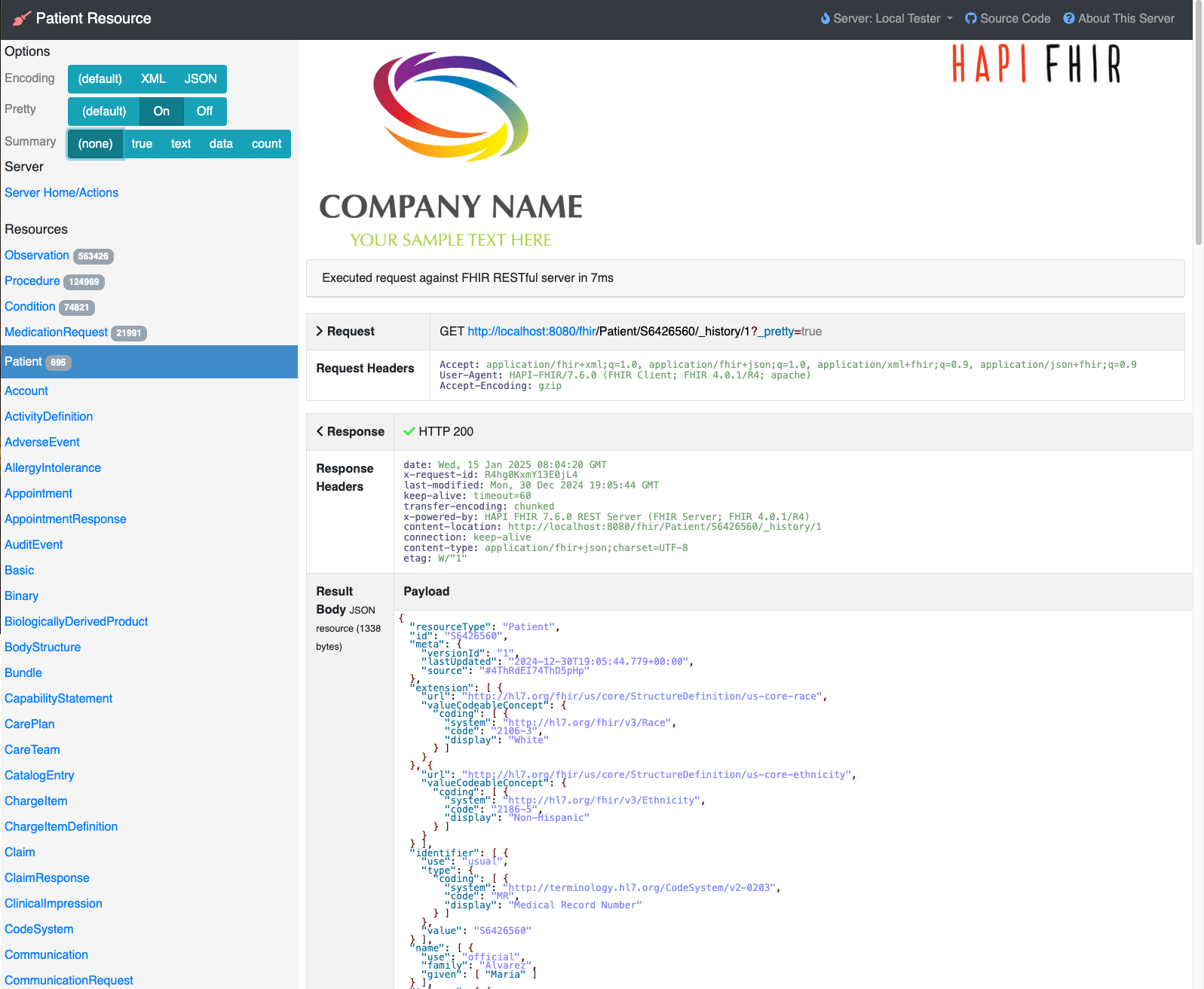}
  \caption{Screenshot of frontend of the FHIR-compliant interactive environment.}
  \label{serverscreen}
\end{figure}

Figure ~\ref{serverscreen} shows a screenshot of frontend of the FHIR-compliant interactive environment.

\subsection{Prompts for the agent system}
Here is the specific prompt used:
\begin{verbatim}
You are an expert in using FHIR functions to assist medical professionals. You are given a
question and a set of possible functions. Based on the question, you will need to make one 
or more function/tool calls to achieve the purpose.

1. If you decide to invoke a GET function, you MUST put it in the format of
GET url?param_name1=param_value1&param_name2=param_value2...

2. If you decide to invoke a POST function, you MUST put it in the format of
POST url
[your payload data in JSON format]

3. If you have answered all the questions and finished all the requested tasks, you MUST 
put it in the format of
finish([answer1, answer2, ...])

Your response must be in the format of one of the three cases, and you SHOULD NOT 
include any other text in the response.

Here is a list of functions in JSON format that you can invoke. Note that you 
should use {api_base} as the api_base.
{functions}

Context: {context}
Question: {question}
\end{verbatim}

\subsection{Subgroup analysis based on difficulty level}

We further break the tasks into three difficulty levels: easy (requires only one step), medium (requires two steps) and hard (requires at least three steps). Table~\ref{diff} in the appendix shows a breakdown of performance on different difficulty levels.

\begin{table}[h]
  \caption{\textbf{Success rate (SR) of state-of-the-art LLMs on MedAgentBench by difficulty levels.} This table presents the success rates of various large language models (LLMs) on MedAgentBench tasks categorized into three difficulty levels: easy (1 step), medium (2 steps), and hard ($\geq$ 3 steps). The highest success rate in each column is highlighted in bold.}
  \centering
  \begin{tabular}{lllllll}
    \toprule
    Model & Size & Form & Overall SR & Easy SR & Medium SR & Hard SR\\
    \midrule
    Claude 3.5 Sonnet v2 & N/A & API &      \textbf{69.67\%} & \textbf{100.00\%} & \textbf{81.67\%} & 23.33\% \\
    GPT-4o & N/A & API &                    64.00\% & 86.67\% & 70.00\% & 33.33\% \\
    DeepSeek-V3 & 685B & open &             62.67\% & 93.33\% & 68.33\% & 24.44\% \\
    Gemini-1.5 Pro & N/A & API &            62.00\% & 82.22\% & 45.83\% & \textbf{63.33\%} \\
    GPT-4o-mini & N/A & API &               56.33\% & 91.11\% & 55.83\% & 22.22\% \\
    o3-mini & N/A & API &                   51.67\% & 67.78\% & 65.00\% & 17.78\% \\
    Qwen2.5 & 72B & open &                  51.33\% & 72.22\% & 44.17\% & 40.00\% \\
    Llama 3.3 & 70B & open &                46.33\% & 56.67\% & 38.33\% & 46.67\% \\
    Gemini 2.0 Flash & N/A & API &          38.33\% & 98.89\% & 17.50\% & 5.56\% \\
    Gemma2 & 27B & open &                   19.33\% & 33.33\% & 23.33\% & 0.00\% \\
    Gemini 2.0 Pro & N/A & API &            18.00\% & 27.78\% & 14.17\% & 13.33\% \\
    Mistral v0.3 & 7B & open &              4.00\% & 13.33\% & 0.00\% & 0.00\% \\
    \bottomrule
  \end{tabular}
  \label{diff}
\end{table}

\end{document}